\documentclass[letterpaper, 10 pt, conference]{ieeeconf}  % Comment this line out if you need a4paper
\IEEEoverridecommandlockouts                              % This command is only needed if 
\usepackage{microtype}
\usepackage{graphicx}
\usepackage{subfig}
\usepackage{caption}
\usepackage{booktabs} % for professional tables
\usepackage{arydshln}
\usepackage{enumitem}
\usepackage{tablefootnote}
\usepackage{xspace}
\usepackage{multirow}

\usepackage{hyperref}

\usepackage{amsmath}
\usepackage{amssymb}
\usepackage{mathtools}
\usepackage{amsthm}

\usepackage[capitalize,noabbrev]{cleveref}

\newcommand{\fig}[1]{Fig. \ref{fig:#1}}
\newcommand{\tabref}[1]{Table \ref{table:#1}}
\newcommand{\chef}{\textsc{chef}\xspace}
\newcommand{\prg}[1]{\noindent\textbf{#1. }}
\theoremstyle{plain}

\theoremstyle{definition}

\theoremstyle{remark}

\title{\LARGE \bf
Mastering Stacking of Diverse Shapes
with Large-Scale Iterative Reinforcement Learning on Real Robots
}

\author{
Thomas Lampe$^{*}$,
Abbas Abdolmaleki$^{*}$,
Sarah Bechtle$^{*}$,
Sandy H. Huang$^{*}$,
Jost Tobias Springenberg$^{*}$,
\\
Michael Bloesch,
Oliver Groth,
Roland Hafner,
Tim Hertweck,
Michael Neunert,
Markus Wulfmeier,
\\
Jingwei Zhang,
Francesco Nori,
Nicolas Heess,
Martin Riedmiller
\thanks{$^{*}$Primary authors; correspondence: \textsc{thomaslampe@google.com}. All authors affiliated with Google DeepMind, London N1C4AG}%
}

\begin{document}
\maketitle
\thispagestyle{empty}
\pagestyle{empty}

%%%%%%%%%%%%%%%%%%%%%%%%%%%%%%%%%%%%%%%%%%%%%%%%%%%%%%%%%%%%%%%%%%%%%%%%%%%%%%%%%%%%%%%%%%%%%%

\begin{abstract}
Reinforcement learning solely from an agent's self-generated data is often believed to be infeasible for learning on real robots, due to the amount of data needed. However, if done right, agents learning from real data can be surprisingly efficient through re-using previously collected sub-optimal data. In this paper we demonstrate how the increased understanding of off-policy learning methods and their embedding in an iterative online/offline scheme (``collect and infer'') can drastically improve data-efficiency by using all the collected experience, which empowers learning from real robot experience only. Moreover, the resulting policy improves significantly over the state of the art on a recently proposed real robot manipulation benchmark. Our approach learns end-to-end, directly from pixels, and does not rely on additional human domain knowledge such as a simulator or demonstrations.
\end{abstract}

%%%%%%%%%%%%%%%%%%%%%%%%%%%%%%%%%%%%%%%%%%%%%%%%%%%%%%%%%%%%%%%%%%%%%%%%%%%%%%%%%%%%%%%%%%%%%%

\section{Introduction}
Recent years have seen significant progress in learning based approaches for the control of real robots. Notably, reinforcement learning has been used to produce high-quality controllers in simulation that can then be transferred to the real world with sim2real approaches \cite{zhu2018reinforcement,jeong2020self,hermann2020adaptive, lee2021beyond,haarnoja2023learning,bohez2022imitate,smith2023learning}, while learning from demonstrations in the form of teleoperated robot trajectories (behavior cloning) directly in the real world has been shown to be surprisingly data-efficient and effective \cite{brohan2022rt,brohan2023rt,stone2023open,bousmalis2023robocat}.

In contrast, reinforcement learning (RL) directly on real robots has received comparatively less attention, despite several attractive properties. Unlike behavior cloning (BC), RL does not rely on expert data. Therefore it is not bounded by the performance of a human teleoperator, does not require potentially expensive teleoperation rigs, and is thus also applicable to robots that cannot be effectively teleoperated. In addition, unlike sim2real, learning directly on real robots does not require a simulator that is carefully matched to reality. Thus it can directly take advantage of the diversity of the real world, and, for instance, of sensors that are available on real robots but hard to simulate.

There are several reasons why RL on real robots has received less attention in recent years. These include questions of data-efficiency, the difficulty of establishing a safe and effective data collection scheme, and also the difficulty of algorithm tuning and of making suitable hyper-parameter choices that avoid instabilities or premature convergence, especially in an online setting.

However, RL algorithms have improved considerably during the last decade. In particular, modern variants of off-policy and offline algorithms are considerably more data-efficient and less sensitive to parameter choices than their predecessors. Importantly, they also allow for a variety of learning scenarios where data collection and policy optimization are interleaved in flexible ways. This makes it possible, for instance, to start and stop data collection as needed, to reuse data from prior experiments and mix different data sources (like self-generated and expert data, or data from different policies), and also to test different algorithm variants and hyper-parameter choices with the same set of data.

This flexibility and its implications have been highlighted by the “Collect-and-Infer” paradigm~\cite{riedmiller2021}, which emphasizes the idea that data collection and policy optimization are two distinct processes that can be optimized separately. In this paper we take inspiration from Collect and Infer, and explore how the aforementioned flexibility can be used to create practical and surprisingly robust and data-efficient iterative schemes for real-robot RL.

As a test case we consider RGB Stacking~\cite{lee2021beyond}, a robot manipulation benchmark that involves stacking of geometric shapes, which emphasizes precision and the understanding of geometric affordances. The contact dynamics of this task are non-trivial to replicate in simulation, thus making it an interesting test case for real robot learning.

Our approach consists of two pairs of online/offline stages, where we collect data during the online phase and subsequently perform offline policy optimization. In the first iteration we (1) perform online off-policy RL to learn an initial policy and collect a diverse dataset. This is followed by (2) a first round of offline policy optimization during which we explore different algorithm settings and model architectures.
(3) In the online phase of the second iteration we evaluate the policies from (2), which sheds light on suitable hyperparameter settings and collects additional, higher-quality data. (4) The final policy is then obtained through offline policy optimization using the full dataset collected so far.

This scheme provides significant flexibility in each phase. For instance, this allows multi-task training for exploration, efficient exploration of different algorithm parameters, and the ability to restart learning with a different network architecture, thus avoiding premature convergence. Overall, this results in a robust and data-efficient learning scheme. Furthermore, training directly on real data removes the need for expensive tuning of a simulator, and allows the policy to take advantage of sensors not available in simulation. The final policy solves the task near flawlessly, outperforming previously published results on this benchmark by a large margin.

%%%%%%%%%%%%%%%%%%%%%%%%%%%%%%%%%%%%%%%%%%%%%%%%%%%%%%%%%%%%%%%%%%%%%%%%%%%%%%%%%%%%%%%%%%%%%%
\section{Background}

\subsection{Markov Decision Processes}
We consider the problem setting of a Markov Decision Process (MDP), defined by a state space $\mathcal{S}$, action space $\mathcal{A}$, transition model $\mathcal{P}: \mathcal{S} \times \mathcal{A} \times \mathcal{S} \rightarrow \mathbb{R}^+$, reward function $r: \mathcal{S} \times \mathcal{A} \rightarrow \mathbb{R}$, discount factor $\gamma \in [0, 1)$, and initial state distribution $\mu_0$. A policy $\pi_\theta(a|s, k)$ specifies a distribution over actions $a \in \mathcal{A}$, given a state $s \in \mathcal{S}$ and task identifier $k \in [1, N]$. The action-value function, or Q-function, for a policy $\pi_\theta$ is the expected return from taking action $a$ in state $s$ and then following this policy: $Q^\pi(s,a; k) = \mathbb{E}_{s' \sim \mathcal{P}(s,a)}[r_k(s,a) + \mathbb{E}_{a' \sim \pi(s')}[Q^\pi(s',a'; k)]]$.

\subsection{Multi-task RL and Scheduled Auxiliary Control (SAC-X)}
\label{sec:sacx}
In the multi-task RL setting, there are $N$ tasks, each of which has a corresponding reward function $r_k$. In this work, we use multi-task RL for both the initial data collection phase and for offline RL, to gather more diverse data and stabilize learning, respectively.

In particular, we use Scheduled Auxiliary Control (SAC-X) \cite{riedmiller2018learning}, which is a multi-task off-policy actor-critic algorithm. The policy and Q-function networks each have a shared torso across tasks, with a separate output head per task. SAC-X is a hierarchical agent, where data is gathered by a scheduler choosing which of the task policies to execute.
In the SAC-Q variant, the scheduler chooses tasks to execute in order to maximize the reward of a predefined main task.

SAC-X trains policies with multi-task policy iteration, alternating between policy evaluation (i.e., updating the Q-functions) and policy improvement (i.e., updating the policies). In the policy evaluation step, for each task $k$, given a batch of transitions $\{s, a, s', \{r_k\}_{k=1}^N\}$ and the current per-task policy $\pi_k^\text{old}$, any policy evaluation algorithm can be used to update the corresponding per-task Q-function $Q_k$. In this work, we use n-step return combined with distributional Q-learning~\cite{bellemare2017}.

For the policy improvement step, SAC-X uses either Stochastic Value Gradients~\cite{heess2015} or maximum a posteriori policy optimization (MPO)~\cite{abdolmaleki2018}. %, adapted for the multi-task setting.
In this work we use MPO for this step, although in theory any policy improvement algorithm could be used, e.g. Soft Actor Critic~\cite{haarnoja2018}.

% Formally, we optimize the multi-task policies by maximizing the MPO objective over all tasks given as:
% \begin{equation}
%     \mathcal{J}(\theta) = \EXP_{\substack{s \sim \mathcal{D} \\ a \sim \pi^\text{old}}} \left[ \sum_{k=1}^N \exp \left(\tfrac{1}{\eta} Q_{\phi}(s, a; k) - Z\right) \log \pi_\theta(a | s, k) \right]\!,
%     \label{eq:eqw}
% \end{equation}
% where the normalizing constant is calculated over a set of sampled actions $Z = \mathbb{E}_{a_1, \dots, a_M \sim \pi^\text{old}}[\log \sum_{i=1}^M \exp(\frac{1}{\eta} Q_{\phi_k}(s, a, k))]$.
% And we learn the corresponding per-task Q-functions by minimizing the following objective wrt. $\phi$,
% \begin{equation}
%     \EXP_{\substack{s, a, s' \\ \in \mathcal{D}}} \left[ \sum_{k=1}^N D\left( Q_\phi(s, a; k) \| r_k + \gamma \EXP_{a' \sim \pi^\text{old}}[Q^\text{old}(s',a'; k)] \right)  \right],
% \end{equation}
% where $D$ denotes that instead of minimizing a squared td-error we are minimizing a distributional error \cite{bellemare2017}, and where $\pi^\text{old}$ and $Q^\text{old}$ denote target networks that use parameters periodically copied (every $200$ updates) from $\theta$ and $\phi$.

\section{Method}
\label{sec:stages}

Our approach, dubbed \chef, is inspired by the Collect and Infer paradigm introduced by \cite{riedmiller2021}. It consists of four stages:
\begin{enumerate}
    \item \textbf{C}ollect a real-world dataset
    \item \textbf{H}yperparameter exploration: offline RL to train policies
    \item \textbf{E}xecute policies on real robots, and collect this data
    \item \textbf{F}ine-tune policies on all collected data
\end{enumerate}

% Aliases here so we can change them all at once if needed. Keep in sync with enumeration above.
\newcommand{\phasecollect}{\textsc{Collect}\xspace}
\newcommand{\phasehyper}{\textsc{Hyperparameter exploration}\xspace}
\newcommand{\phaseeval}{\textsc{Execute}\xspace}
\newcommand{\phasefinetune}{\textsc{Fine-tune}\xspace}

\begin{figure}[t]
    \centering
    \includegraphics[width=0.8\columnwidth]{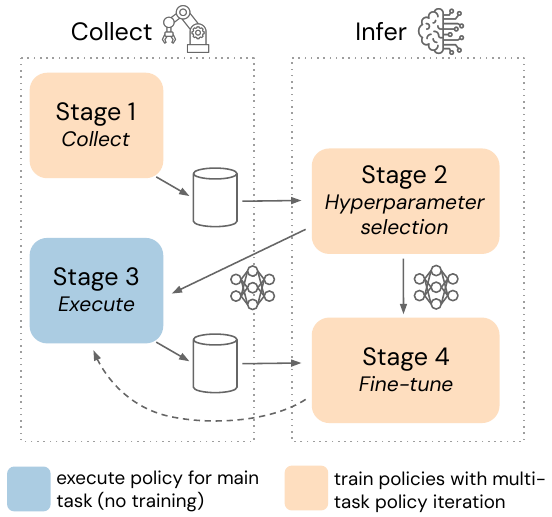}
    \caption{Illustration of our approach, \chef, and how it relates to the more general collect-and-infer paradigm. We collect data in Stages 1 and 3, and train policies with offline RL on the data in Stages 2 and 4.}
    \label{fig:collect_and_infer}
    \vspace{-1.5em}
\end{figure}

In this framework, no data is wasted---in Stages 1 and 3, we collect real-world data, while in Stages 2 and 4, we train policies on the collected data. No new data is collected in Stages 2 and 4. In Stage 3, we evaluate policies with no additional training; this is necessary regardless for evaluating the performance of policies trained fully offline.
% Importantly, we can repeat Stages 3 and 4 until the desired performance is obtained, while accumulating the collected data -- although this was not necessary for the results presented here. Such \emph{iterative collect and infer} allows our algorithm to 'hill-climb' to high performance via offline-RL starting from initial data.

\prg{\phasecollect (Stage 1)}
We use SAC-Q to train multi-task policies from scratch on real robots. The SAC-Q algorithm uses the scheduler to decide which sequence of task policies to execute in each episode. By optimizing policies for different tasks and sequencing them in multiple ways, this algorithm achieves better coverage of the state space compared to single task training, and thus collects to a more diverse dataset. We continue Stage 1 until the policies for all tasks have converged (albeit to a sub-optimal solution), and store all interaction data for future use.

\prg{\phasehyper (Stage 2)}
In Stage 2, we train agents in an offline RL setting where the transitions come from the dataset collected in Stage 1. We collect no new data in this phase. We sweep over a variety of hyperparameters, for instance network architecture and training algorithm parameters. We train agents with multi-task policy iteration; this is exactly the same as in Stage 1, except now the transitions come from a fixed dataset rather than from executing the scheduler on a real robot. This works surprisingly well, despite observations made elsewhere that typical off-policy RL algorithms cannot be directly applied in the offline RL setting, due to extrapolation error~\cite{fujimoto2019}. In contrast we here find that multi-task training with a suitable architecture sufficiently stabilizes offline learning if data-coverage of the state-action space is good (as promoted by our multi-task data collection).

\prg{\phaseeval (Stage 3)}
We execute fully-trained policies from each of the hyperparameter choices in Stage 2, on real robots, and save all the data gathered. This allows us not only to determine which of the hyperparameters results in the best policy, but also to gather extra high-quality data. This minimizes the additional data necessary for hyperparameter selection. In contrast, na\"{\i}ve hyperparameter selection would require either training from scratch for each choice of hyperparameters, or selecting hyperparameters based on what works best in simulation, and hoping that those work on the real system.

\prg{\phasefinetune (Stage 4)}
We then use all of the data gathered so far (from Stages 1 and 3) to fine-tune the best policies offline, based on the real-world evaluations from Stage 3.

%%%%%%%%%%%%%%%%%%%%%%%%%%%%%%%%%%%%%%%%%%%%%%%%%%%%%%%%%%%%%%%%%%%%%%%%%%%%%%%%%%%%%%%%%%%%%%

\section{Related Work}

\subsection{Offline RL for robot learning}

Offline RL is the data-driven formulation of the reinforcement learning problem. The aim is still to maximize reward; however, the agent can no longer interact with the environment and collect additional
transitions \cite{levine2020offline}. Recent actor-critic algorithms for offline RL include Critic Regularized Regression \cite{wang2020critic} for simulated continuous control tasks; conservative Q-learning \cite{kumar2020conservative}, which learns a lower bound on the policy value for stabilizing offline RL; and QT-Opt \cite{kalashnikov2018qt}, a self-supervised vision-based framework. More recently in \cite{chebotar2021actionable}, goal-conditioned offline Q-learning is used to learn robotic skills of discrete actions from previously collected offline data without access to specified rewards; this enables learning a variety of skills on real robots. Beyond actor-critic algorithms, in \cite{cabi2019scaling} a reward model is learned from human preferences to annotate existing offline datasets, which is then used to perform batch RL \cite{Lange2012}. In \cite{lee2022spend} a teacher student learning setup is presented where a dataset collected by a suboptimal teacher is used for offline RL to warm-start the student policy. In this work we present an offline RL algorithm that learns an actor and critic in a multi-task fashion, alternating between online data collection and offline learning to master the RGB Stacking task from \cite{lee2021beyond} and used in \cite{lee2022spend}.

\subsection{Multi-task RL for robot learning}

Multi-task RL holds the promise of amortizing the cost of single task learning by providing a shared representation across tasks. An approach to multi-task RL is to condition policies on tasks  \cite{deisenroth2014multi, riedmiller2018learning, wulfmeier2020compositional} or to distill single task policies into a shared multi-task policy \cite{teh2017distral, rusu2015policy, levine2016end, parisotto2015actor,ghosh2017divide}.

Other works create a mapping between tasks and individual policy parameters, in order to select the adequate policy for the specific task at hand \cite{kober2012reinforcement, da2012learning}. All these approaches present fairly hierarchical setups to solving the multi-task RL problem. In \cite{kalashnikov2021scaling} the authors show that scaling up multi task RL with real robot data enables transfer and even zero-shot generalization.
Similarly, this work focuses also on using real robot data for actor-critic learning; however, our work aims to minimize the amount of real robot data needed, by training policies with offline multi-task RL.

%%%%%%%%%%%%%%%%%%%%%%%%%%%%%%%%%%%%%%%%%%%%%%%%%%%%%%%%%%%%%%%%%%%%%%%%%%%%%%%%%%%%%%%%%%%%%%

\section{Benchmark Description}

Block stacking has long been a standard benchmark task for robotic manipulation, with early work on vision-based block stacking from~\cite{deisenroth2014multi}. Recent works have tackled this problem by learning a curriculum of the different stages of the task \cite{riedmiller2018learning}, combining human demonstrations and RL \cite{zhu2018reinforcement,cabi2019scaling}, enabling sim-to-real transfer \cite{zhu2018reinforcement,jeong2020self,hermann2020adaptive, lee2021beyond}, and learning a generalist transformer-based policy from expert demonstrations  \cite{bousmalis2023robocat}.

The RGB Stacking benchmark we consider in this work was first presented in \cite{lee2021beyond, lee2022spend}.
The task setup consists of 5 distinct block stacking configurations, that consist of parameterized geometric shapes of red, green and blue objects.
This is a challenging benchmark for robotic manipulation. In contrast to our work that learns to master stacking purely from data collected on the real robot, the authors in \cite{lee2021beyond} adopt a sim-to-real approach with a final fine-tuning stage on hardware.
The manipulator used in this work is the Rethink Sawyer robot \cite{sawyer}, which is controlled here in task space, with a 5-dimensional action consisting of the end effector's translational velocity in Cartesian space, the wrist rotation velocity, and the velocity of the arm's parallel gripper.
The policy has access to a mixture of proprioceptive observations (positions, velocities and forces of the end effector and joints) and 3 low-resolution (128x128) images from stationary cameras attached to the workspace.
The observations notably exclude the positions of the objects; while those are tracked for the purpose of automation and reward computation, the agent must learn to act from vision alone.

\begin{figure*}
  \centering
  \begin{minipage}[t]{.38\linewidth}
    \subfloat[Robot setups used in the RGB Stacking benchmark, showing the Sawyer arm, basket, and attached cameras.
        \label{fig:robot_cell}
    ]{
        \includegraphics[width=\linewidth]{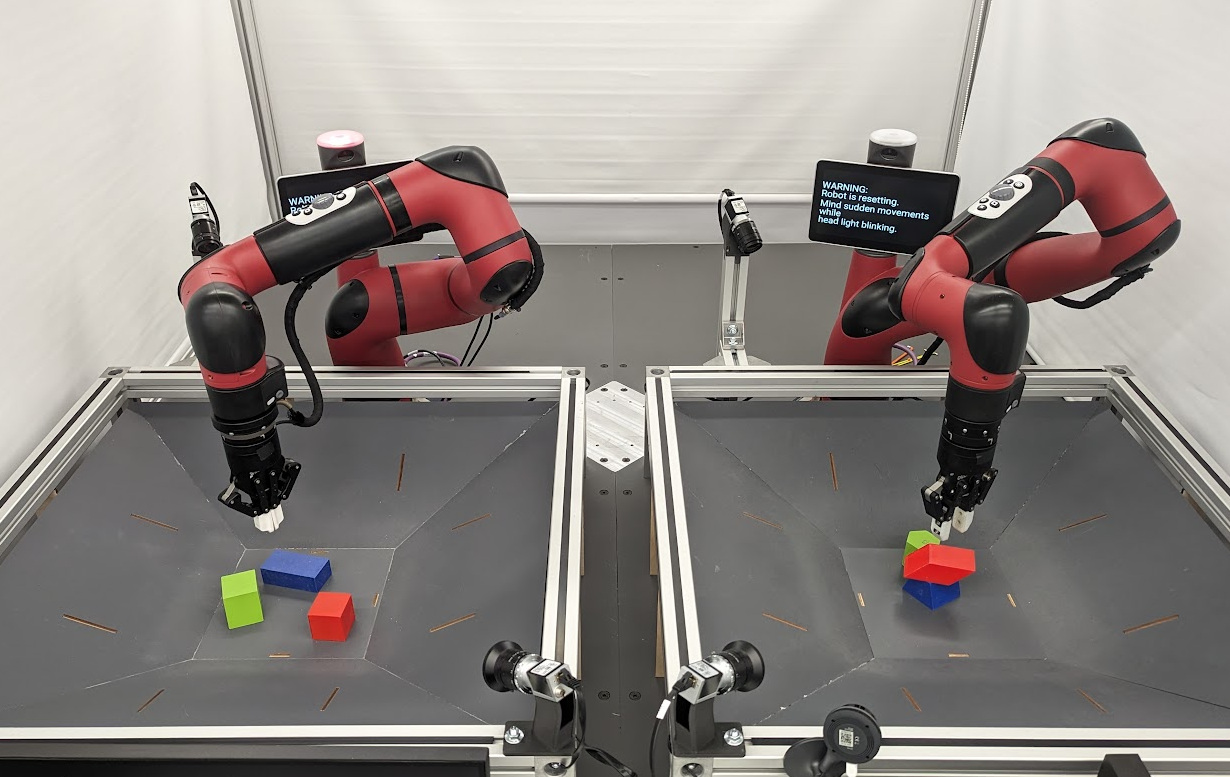}
    }%
  \end{minipage}%
  \hfill
  \begin{minipage}[t]{.205\linewidth}
    \subfloat[Triplets 1 to 5 (top to bottom) in the skill mastery challenge.
        \label{fig:triplets}
    ]{
        \includegraphics[width=\linewidth]{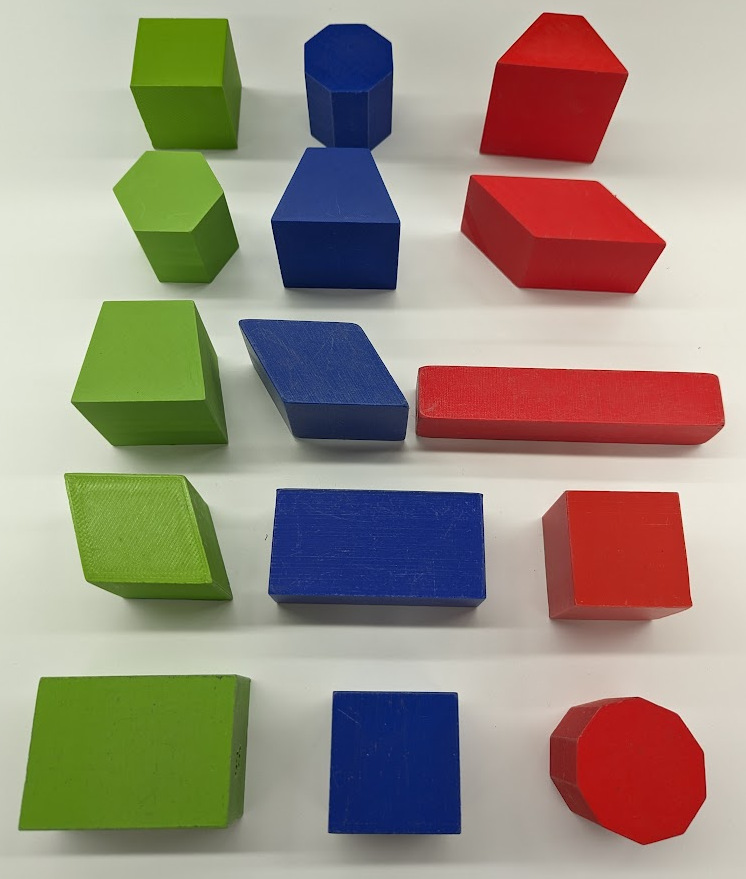}
    }%
  \end{minipage}%
  \hfill
  \begin{minipage}[t]{.33\linewidth}
    \subfloat[Cropped, scaled images given to the agent.
        \label{fig:agent_view}
    ]{
        \includegraphics[height=.3\linewidth]{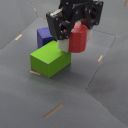} \\
        \includegraphics[height=.3\linewidth]{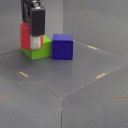} \\
        \includegraphics[height=.3\linewidth]{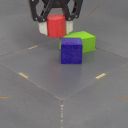} \\
    }\\
    \subfloat[Average \textbf{stack-leave} success in \phasecollect.
        \label{fig:online_performance}
    ]{
        \includegraphics[width=0.9\linewidth]{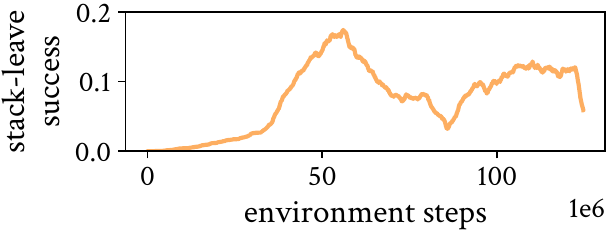}
    }%
  \end{minipage}%
  \caption{
      The robot, objects, and agent inputs comprising the RGB Stacking benchmark, and learning in the \phasecollect stage.
  }
  \vspace{-0.5em}
  \label{fig:rgb_stacking} 
\end{figure*}

The role of the shapes is color-coded: the red object should be stacked on the blue one, and a third green one serves as a distractor.
Due to the different shapes of the objects, an agent is forced to learn about various geometric properties, including slanted faces, lengths, and off-center balancing.

This task remains challenging for existing methods to master, especially because success depends both on the contact physics of objects and on force interactions, e.g. when objects are not oriented in a way that would enable stacking, and need to be carefully nudged onto their side.
Both aspects can prove difficult to model in simulation with sufficient fidelity to enable transfer, and are also hard to demonstrate via teleoperation due to limited force feedback.

In this work, we are specifically considering the ``skill mastery'' sub-task, where the goal is to achieve maximum performance with a single policy on five test object triplets (\fig{triplets}), and the test objects are available during training.

\section{Experiment Setup}

\subsection{Sub-tasks}

The SAC-Q multi-task learning algorithm described in Section \ref{sec:sacx} requires defining a set of sub-tasks.
For the sub-task rewards we use the components of the composite reward in \cite{lee2021beyond}.
These form a natural curriculum from simpler to more complex multi-stage behaviors, to achieve the main \textbf{stack-leave} task.
We describe them conceptually below; for a full description including mathematical formulations, see~\cite{lee2021beyond}.
\begin{description}
    \item[open] Proportional to the opening angle of the gripper.
    \item[reach-grasp] Sum of a shaped component for moving close to the red object, and a sparse component for triggering the gripper's contact detection.
    \item[lift] Shaped reward, proportional to the red object's height.
    \item[place] Shaped reward, decreases non-linearly as the red object approaches a position directly above the blue object.
    \item[stack] Sparse reward, non-zero when the red object is on top of the blue object, with a small tolerance.
    \item[stack-leave] Sparse reward, non-zero when the conditions for \textbf{stack} hold, and the tool center point of the gripper is at least 10$cm$ from the red object's center of mass.
\end{description}

\subsection{\chef Stages}
As per the \chef approach in Section \ref{sec:stages}, we perform training in distinct stages, concretely performed as follows:

\begin{description}
    \item[\phasecollect:] Online RL for approximately 330k episodes. This stage is distributed across a fleet of identical setups, with 10 robots collecting episodes, writing their data to a shared experience replay, and retrieving the newest online-trained policy after each episode. We train the policies until the task return converges (\fig{online_performance}). 
    \item[\phasehyper:] We vary several hyperparameters, described in Section \ref{sec:results}. Candidate policies are trained for 2M update steps each, at which point performance has reliably converged.
    \item[\phaseeval:] Evaluating the policies generated in the previous stage results in approximately 70k additional episodes. All of these evaluations are performed for the \textbf{stack-leave} sub-task only, even for multi-task policies.
    \item[\phasefinetune:] We continue training the best policy from the previous stage for another 2M steps, using the combined 400k episodes from both data collection phases.
\end{description}

\subsection{Model Architecture}

For both the critic and the actor network, we use a Residual Neural Network~\cite{he2015deep} (ResNet) for the shared torso, with a separate MLP head per task.
Details are provided in the supplementary material\footnote{\url{https://sites.google.com/view/robochef}}.
Most parameter settings are shared across all experiments, and are chosen based on prior simulation experiments.
The most influential parameter we vary is the number and size of ResNet channels; these will be compared in the results below.

\subsection{Evaluation}

In the \phaseeval stage, we evaluate each candidate policy for 200 episodes.
Between episodes, a hand-written controller randomizes the positions of all objects in the workspace, and moves the end effector to a random position.

As a performance metric, we report the percentage of episodes deemed ``successful'', defined by the \textbf{stack-leave} reward component being 1 at the end of the episode.
Each episode lasts for 20 seconds, at a control rate of 20 Hz.
Episodes are terminated prematurely if the wrist force-torque sensor registers a force above 20$N$ at any time.

%%%%%%%%%%%%%%%%%%%%%%%%%%%%%%%%%%%%%%%%%%%%%%%%%%%%%%%%%%%%%%%%%%%%%%%%%%%%%%%%%%%%%%%%%%%%%%

\section{Results}
\label{sec:results}

We first present the final performance of the policy trained through our approach, compared to state-of-the-art as baselines (\tabref{finalperformance}).
A single iteration of offline RL is in fact already sufficient to outperform the baselines ($92\%$ vs $82\%$ for the strongest baseline). This is a remarkably large improvement, compared to the policy at the end of the \phasecollect stage ($26\%$). We attribute this large jump to the fact that the algorithm now has access to the full experience from the start of training. Additionally, no care needs to be taken to avoid overfitting when the data distribution is still narrow, so the size of the model can be increased. Nonetheless, performance after this first offline phase is not optimal. It is only after the second round of collect-and-infer that performance reaches near-optimality ($96\%$).

\begin{table}[t]
    \centering
    \footnotesize
    \begin{tabular}{l@{\hspace{1.5\tabcolsep}} c@{\hspace{1\tabcolsep}}c@{\hspace{1\tabcolsep}}c@{\hspace{1\tabcolsep}}c@{\hspace{1\tabcolsep}}c@{\hspace{1\tabcolsep}} c}
    \toprule
    Method & \multicolumn{5}{c}{Triplet} & Average \\
           & 1 & 2 & 3 & 4 & 5 & \\
    \midrule
    BC (S2R + R) \cite{lee2021beyond}  & 76\% & 61\% & 71\% & 88\% & 78\% & 75\% \\
    CRR (S2R + R) \cite{lee2021beyond} & 87\% & 68\% & 75\% & 88\% & 89\% & 82\% \\
    R-MPO (S2R + R) \cite{lee2022spend}     & 82\% & 55\% & 73\% & 92\% & 89\% & 78\% \\
    R-CRR (S2R + R) \cite{lee2022spend}     & 54\% & 61\% & 74\% & 94\% & 94\% & 75\% \\
    \midrule
    \phasecollect  &  6\% & 22\% & 27\% & 47\% & 29\% & 26\% \\
    \textsc{Hyperparam.}    & 92\% & 89\% & 93\% & 92\% & 97\% & 92\% \\
    \phaseeval     & 62\% & 50\% & 63\% & 74\% & 71\% & 61\% \\
    \phasefinetune & \textbf{96\%} & \textbf{97\%} & \textbf{95\%} & \textbf{95\%} & \textbf{98\%} & \textbf{96\%} \\
    \bottomrule
    \end{tabular}
    \caption{
        Comparison between prior work and the four \chef stages.
        Performance for \phasehyper is for the best policy, chosen for fine-tuning later.
        Performance for \phaseeval is the average of all evaluations. For baselines, S2R denotes sim-to-real R denotes real-data used.
    }
    \vspace{-1.5em}
    \label{table:finalperformance}
\end{table}

In the remainder of this section, we will discuss the three design choices we investigated in the \phasehyper stage: inclusion of real-world-only observations, multi-task offline learning, and network size.
% The best policy from the \phasehyper stage incorporates three main design choices: an optimized observation selection, multi-task offline learning, and an increased network size compared to in the \phasecollect stage.
% Each of these choices will be discussed in the following subsections.
We will also present several ablations, on how the final \phasefinetune policy is trained and how the \phasecollect and \phasehyper stages are designed.

\begin{figure*}[t]
    \centering
    \includegraphics[width=0.27\columnwidth]{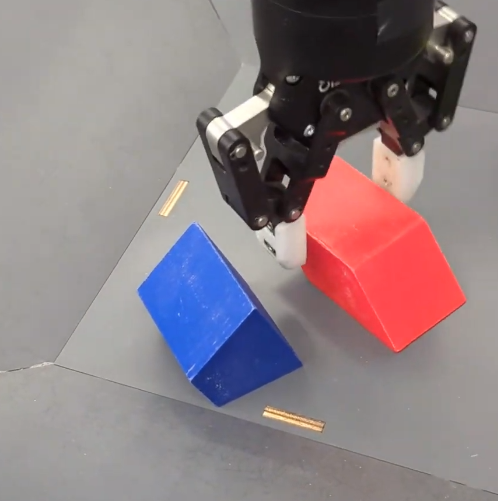}
    \includegraphics[width=0.27\columnwidth]{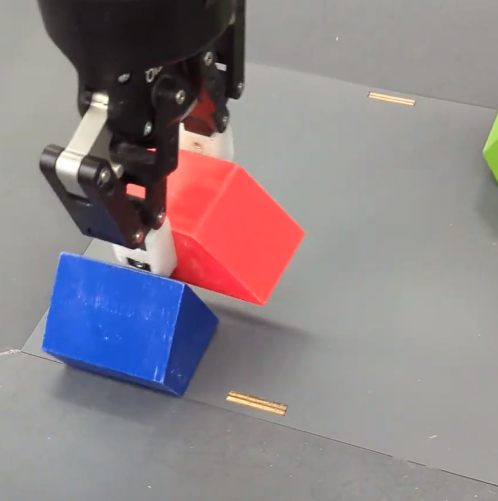}
    \includegraphics[width=0.27\columnwidth]{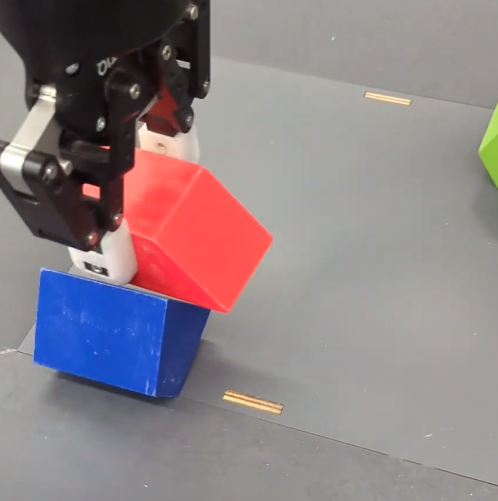}
    \includegraphics[width=0.27\columnwidth]{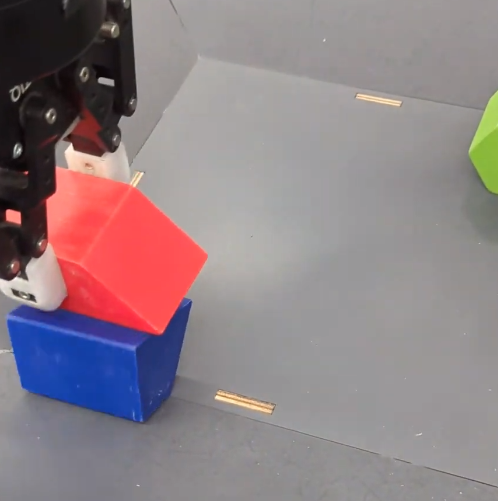}
    \includegraphics[width=0.27\columnwidth]{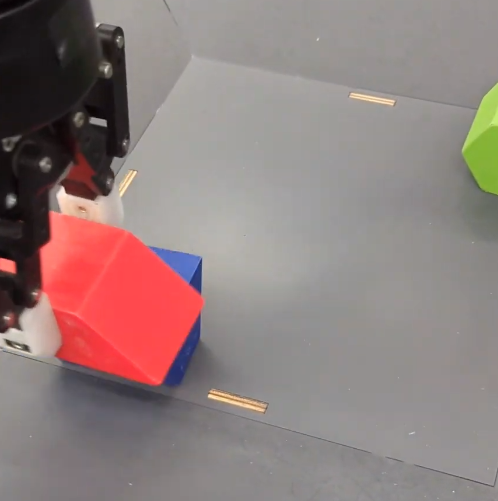}
    \includegraphics[width=0.27\columnwidth]{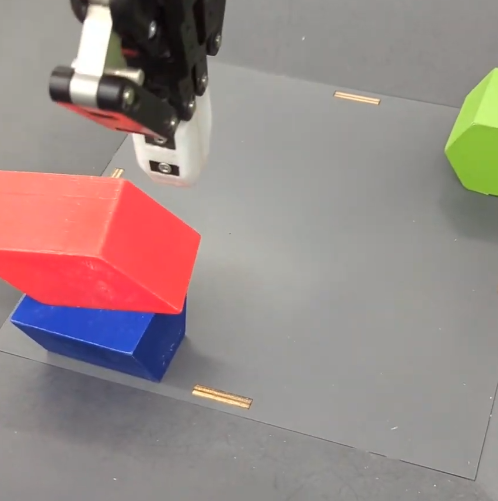}
    \caption{Successful stacking attempt for triplet 2. The agent first aligns the blue object with the basket's edge and then carefully pushes against it in order to flip it into an orientation that the red object can be stacked onto.}
    \vspace{-0.5em}
    \label{fig:triplet2}
\end{figure*}

\subsection{Hyperparameter Exploration: Observations}
\label{sec:sim2real}

One of the main motivations for learning purely from real data, rather than also leveraging simulators, is the ability to rely on sensor data that is difficult to simulate.
In particular, good performance on some test triplets involves gentle interaction with the objects.
For Triplet 2, the blue object may need to be flipped over in order to stack on top of it, as illustrated in \fig{triplet2}.
This must be done without exceeding wrist force-torque thresholds, which would trigger an episode termination.
That in turn requires the use of the robot's wrist force sensor, which was left out in previous work because the large sim-to-real gap of these measurements was found to hinder transfer.

Using only simulation-capable observations leads to an average success rate of $43\%$ on the relevant Triplet 2.
In contrast, including haptic observations, namely forces and torques for both wrist and joint sensors, significantly increases the success rate to $78\%$ after the first round of offline RL, with all other settings being equal.
Qualitatively, this leads to more deliberate policies that carefully nudge the bottom object when necessary, and reliably avoid force-torque-based early terminations.
For Triplet 2, the percentage of episodes terminated early drops from $~30.2\%$ without haptic observations, to $~5.6\%$ when including them.
Examples of this behaviour are provided in the supplementary video.

\subsection{Hyperparameter Exploration: Network Size}
\label{sec:networksize}

An advantage of offline learning is that it enables us to switch the network architecture after the initial \phasecollect phase. For collection, it can be preferable to use a smaller model, to reduce the risk of overfitting and increase the speed of parameter updates: we used a ResNet with channel sizes of $\{16, 32, 32\}$ and an embedding size of 32.
However, for the final offline policy, a larger model is preferable, to ensure that the policy can be fit with maximum precision.
Increasing the channel sizes to $\{64, 128, 128, 64\}$ and the embedding size to 256 yields a higher success rate (\tabref{ablations}-B). 

\subsection{Hyperparameter Exploration: Multi-task Offline Learning}
\label{sec:bc}
\label{sec:multitask_real}

Despite the previously described advantages of offline RL, it can be unstable~\cite{fujimoto2019}.
Behavioral cloning (BC) provides a mechanism for stabilizing learning, but its performance is limited in our setup, because the \phasecollect dataset contains many non-successful episodes.
If we filter the data to include only successful episodes, BC still suffers from poor data coverage.
In contrast, offline RL can utilize the entire dataset, and obtains higher performance than BC (\tabref{ablations}-C).

The better performance of our offline RL approach hinges on the use of multi-task RL, as well as non-expert data and the overall data distribution \cite{lambert2022challenges}.
We use an SAC-Q multi-task critic as described in Section \ref{sec:sacx}, with the same six sub-tasks as used during the initial \phasecollect stage.
For comparison, we ran single-task offline RL for only the \textbf{stack-leave} reward, and this performs significantly worse.
Likewise, multi-task offline RL struggles to learn with the same success-filtered dataset as used for BC, as the critic overestimates the value of non-covered states and actions.

\begin{table}
    \centering
    \footnotesize
    \begin{tabular}{l@{\hspace{1.5\tabcolsep}}l@{\hspace{1.5\tabcolsep}} c@{\hspace{1\tabcolsep}}c@{\hspace{1\tabcolsep}}c@{\hspace{1\tabcolsep}}c@{\hspace{1\tabcolsep}}c@{\hspace{1\tabcolsep}} c}
    \toprule
    & Method & \multicolumn{5}{c}{Triplet} & Average \\
    &        & 1 & 2 & 3 & 4 & 5            & \\
    \midrule
    \multirow{2}{2.5em}{\ref{sec:networksize}} & Small network & 87\% & 80\% & 86\% & 89\% & 91\% & 86\% \\
    & Large network & \textbf{92\%} & \textbf{89\%} & \textbf{93\%} & \textbf{92\%} & \textbf{97\%} & \textbf{92\%} \\
    \midrule
    \multirow{4}{2.5em}{\ref{sec:multitask_real}} & Multi-task RL  & \textbf{87\%} & \textbf{80\%} & \textbf{86\%} & 89\% & 91\% & \textbf{86\%} \\
    & Single-task RL & 51\% & 75\% & 60\% & \textbf{90\%} & \textbf{93\%} & 74\% \\
    & Filtered BC    & 78\% & 60\% & 48\% & 80\% & 91\% & 71\% \\
	& Filtered RL    & 1\% & 3\% & 7\% & 27\% & 13\% & 11\% \\
	\midrule
	\multirow{4}{2.5em}{\ref{sec:finetuning}} & Iteration 1      & 91\% & 73\% & 76\% & 92\% & 91\% & 84\% \\
	\cdashline{2-8}[3pt/2pt] \noalign{\vskip 0.5mm}
    & Fine-tuning 2M & \textbf{97\%} &\textbf{90\%} & \textbf{94\%} & \textbf{96\%} & 98\% & \textbf{95\%} \\
    & Re-training 2M & 93\% & 78\% & 91\% & \textbf{96\%} & \textbf{99\%} & 91\% \\
    & Re-training 4M & 96\% & 87\% & 93\% & \textbf{96\%} & 98\% & 94\% \\
    \bottomrule
    \end{tabular}
    \caption{Success rates for ablations of our approach.}
    \label{table:ablations}
    \vspace{-1em}
\end{table}

\subsection{Ablation: Fine-tuning}
\label{sec:finetuning}

For the final stage, we fine-tune the best policy, based on the results from the \phaseeval stage.
Every mini-batch consists equally of samples drawn from the non-expert, multi-task dataset from the \phasecollect stage, and the more success-heavy, single-task dataset produced during the \phaseeval stage.

Compared to training from scratch, fine-tuning reaches similar performance in half as many learner updates---2M as opposed to 4M (\tabref{ablations}-D). When training from scratch for only 2M updates, performance is noticeably lower on the challenging Triplet 2.
To put the reduction of required compute into scale, it takes approximately 18 days to train the large network for 2M learner updates, on 16 Google Cloud TPU v3 accelerators.

%%%%%%%%%%%%%%%%%%%%%%%%%%%%%%%%%%%%%%%%%%%%%%%%%%%%%%%%%%%%%%%%%%%%%%%%%%%%%%%%%%%%%%%%%%%%%%

\begin{figure*}[t]
  \centering
  \begin{minipage}[t]{\columnwidth}
  \subfloat[Online RL, analogous to \phasecollect. The reward for \textbf{stack-leave} across training, for the data collection policy (left) and the task policy (right). Sharing the critic torso improves sample efficiency and final performance. When the critic torso is not shared, there is a delay from when the data collection starts obtaining reward for \textbf{stack-leave} after around 5M environment steps, compared to when the \textbf{stack-leave} policy learns from this data, more than 5M steps later.
    \label{fig:shared_torso}
    ]{
    \includegraphics[width=\columnwidth]{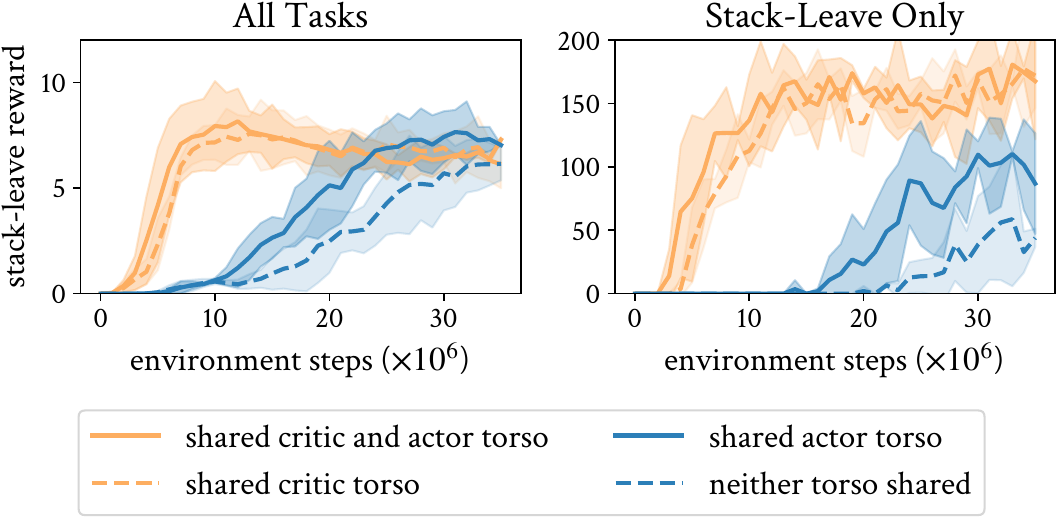}
  }
  \end{minipage}%
  \hfill
  \begin{minipage}[t]{\columnwidth}
    \subfloat[Offline RL, analogous to \phasehyper. Plots show performance of the \textbf{lift} and \textbf{stack} task policies. Sharing the critic network torso across tasks does not impact performance on \textbf{lift}, but is essential for learning \textbf{stack}, a harder and sparse-reward task. The multi-task setup is important for both gathering data (i.e., \phasecollect) and learning from this data, as shown by the worse performance on \textbf{stack} (orange dotted and grey lines, respectively).
    \label{fig:offline_rl}
    ]{
    \includegraphics[width=\columnwidth]{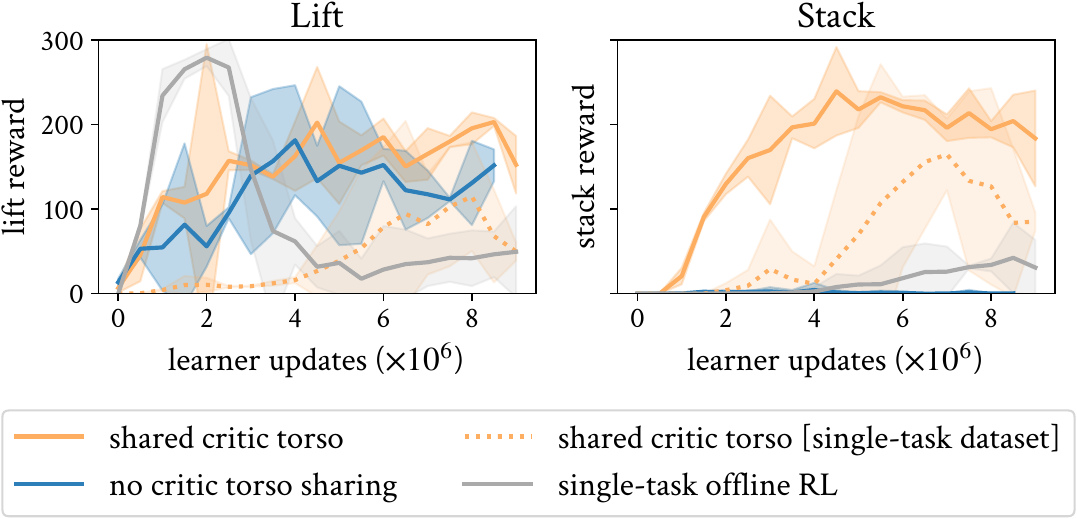}
  }
  \end{minipage}%
  \vspace{-1em}
  \caption{
      Ablations in simulation, on sharing network torsos during online and offline learning, and how the dataset is gathered.
  }
  \vspace{-1.5em}
\end{figure*}

\subsection{Ablations in Simulation}
\label{sec:multitask_sim}

While we did not use simulation for pre-training or transfer, we did rely on it to inform the design of the \phasecollect and \phasehyper stages.

A key component of our approach is sharing the critic network torso across tasks. In simulation, we ablate this for the same RGB stacking benchmark, but with state features instead of image inputs. In the online RL setting, which is analogous to \phasecollect, we found that sharing the critic torso leads to both faster learning and better performance for the main \textbf{stack-leave} task, whereas sharing the actor torso is less crucial (\fig{shared_torso}, right). Note that much of the data in the experience replay is off-policy with respect to an individual task's policy, since during data collection, the scheduler sequences all the task policies together. Sharing the critic torso enables learning from this off-policy data, as shown by the shorter delay between when the data collection policy obtains non-zero reward for \textbf{stack-leave} (\fig{shared_torso}, left) and when the task policy starts learning from this data. We hypothesize that sharing the critic torso enables learning a shared representation that simplifies critic learning for harder, sparse-reward tasks like \textbf{stack-leave}. We found that when the critic torso is not shared, the critic for \textbf{stack-leave} overestimates the return.

In the offline RL setting, analogous to \phasehyper, sharing the critic network torso is even more important. Here we focus on a subset of four tasks:  \textbf{reach-grasp}, \textbf{lift}, \textbf{place}, and \textbf{stack}. We first gather 40k episodes with multi-task online RL, as in \phasecollect. Without a shared critic torso, the agent learns the \textbf{lift} task (\fig{offline_rl}, left) but cannot learn the sparse-reward \textbf{stack} task (\fig{offline_rl}, right).

We also run two other ablations in this offline RL setting. First, we instead gather the dataset with single-task online RL, where the reward is the composite reward used in \cite{lee2021beyond}. Second, instead of using multi-task offline RL, we train a policy to optimize the single composite reward. Both cannot learn to stack consistently. These ablations indicate the importance of using a multi-task setup. Gathering the dataset in a multi-task setup improves coverage and includes trajectories that perform well for each of the individual tasks. Training policies offline in a multi-task setup enables using the learning signal from easier tasks to simplify learning for harder tasks.

%%%%%%%%%%%%%%%%%%%%%%%%%%%%%%%%%%%%%%%%%%%%%%%%%%%%%%%%%%%%%%%%%%%%%%%%%%%%%%%%%%%%%%%%%%%%%%

\section{Discussion}

We presented the \chef schema, a specific implementation of the collect-and-infer paradigm, targeted toward off-policy actor-critic reinforcement learning.
By separating the learning into distinct stages for exploration-driven data collection, offline hyperparameter search, greedy execution, and offline fine-tuning, we achieved maximum reuse of collected data for improved data efficiency, while also maintaining flexibility with regards to architecture and hyperparameters.

We applied this schema to the RGB Stacking benchmark, on real robots.
Our approach not only reduced the required amount of system interaction, but also achieved near-perfect success, significantly improving upon the state-of-the-art.
Our ablations highlighted several key features of the approach:
First, using end-to-end RL to train directly on real robots, without the need for a simulator, allowed the use of relevant sensors that were previously ignored due to limited fidelity in simulation (Section \ref{sec:sim2real}).
Second, offline RL coupled with multi-task exploration was able to utilize the large quantity of non-successful data produced during early training, outperforming filtered BC (Section \ref{sec:bc}). In addition, using a multi-task critic network architecture was key for stabilizing offline RL (Sections \ref{sec:multitask_real}, \ref{sec:multitask_sim}). Finally, by using fine-tuning in the second offline training stage, we achieved similar performance compared to training from scratch, with half the computational cost (Section \ref{sec:finetuning}).

In this work, using a set of sub-tasks that form a curriculum towards the desired main task worked well. In future work, we would like to explore how the selection of tasks affects these findings, in particular as increasingly large models are being used to train many tasks across many domains at once (e.g. \cite{brohan2022rt,brohan2023rt,bousmalis2023robocat}).

To solve the RGB Stacking benchmark task, we only needed to perform the \phaseeval and \phasefinetune stages once. If required, we could also repeat these stages until the desired performance is obtained, while accumulating the collected data. Such \emph{iterative Collect-and-Infer} would allows the algorithm to 'hill-climb' to high performance via offline-RL starting from initial data. Additionally, the question arises whether repeating the \phasehyper stage is optimal, or whether additional exploration would be required. While the answer may depend on the specific task considered, therein lies the strength of the general Collect-and-Infer framework, to choose the specific stages as needed.

%%%%%%%%%%%%%%%%%%%%%%%%%%%%%%%%%%%%%%%%%%%%%%%%%%%%%%%%%%%%%%%%%%%%%%%%%%%%%%%%%%%%%%%%%%%%%%

\bibliography{example_paper}
\bibliographystyle{IEEEtran}

\end{document}